\documentclass[]{fairmeta}
\usepackage{amsmath}    
\usepackage{amssymb}    
\usepackage{mathtools}  
\usepackage{bm}         
\usepackage{algorithm}
\usepackage{algpseudocode}
\usepackage{paralist}
\usepackage[dvipsnames,table,xcdraw]{xcolor}

\definecolor{skyblue}{RGB}{70,130,180} 

\newcommand{\linchange}[1]{\textcolor{black}{#1}}
\newcommand{\linchangenew}[1]{\textcolor{black}{#1}}
\title{Paying Less Generalization Tax: A  Cross-Domain Generalization Study of RL Training  for LLM Agents}
\author[1,3,*,\circ]{Zhihan Liu}
\author[1,*]{Lin Guan}
\author[1]{Yixin Nie}
\author[1,4, \circ]{Kai Zhang}
\author[1,5, \circ]{Zhuoqun Hao}
\author[1]{Lin Chen}
\author[2]{Asli Celikyilmaz  }
\author[3,\dagger]{Zhaoran Wang}
\author[1,\dagger]{Na Zhang}

\affiliation[1]{Meta Superintelligence Labs}
\affiliation[2]{FAIR at Meta}
\affiliation[3]{Northwestern University}
\affiliation[4]{The Ohio State University}
\affiliation[5]{University of Pennsylvania}
\contribution[\circ]{Work done at Meta}
\contribution[*]{Joint First Author}\contribution[\dagger]{Joint Last Author}

\abstract{Generalist LLM agents are often post-trained on a narrow set of environments but deployed across far broader, unseen domains. In this work, we investigate the challenge of agentic post-training when the eventual test domains are unknown. Specifically, we analyze which properties of reinforcement learning (RL) environments and modeling choices have the greatest influence on out-of-domain performance. First, we identify two environment axes that strongly correlate with cross-domain generalization: (i) state information richness, i.e., the amount of information for the agent to process from the state, and (ii) planning complexity, estimated via goal reachability and trajectory length under a base policy. Notably, domain realism and text-level similarity are not the primary factors; for instance, the simple grid-world domain Sokoban leads to even stronger generalization in SciWorld than the more realistic ALFWorld. Motivated by these findings, we further show that increasing state information richness alone can already effectively improve cross-domain robustness. We propose a randomization technique, which is low-overhead and broadly applicable: add small amounts of distractive goal-irrelevant features to the state to make it richer without altering the task.
Beyond environment-side properties, we also examine several modeling choices: (a) SFT warmup or mid-training helps prevent catastrophic forgetting during RL but undermines generalization to domains that are not included in the mid-training datamix; and (b) turning on step-by-step thinking during RL, while not always improving in-domain performance, plays a crucial role in preserving generalization. Based on these findings, we offer practical guidance for agentic post-training when deployment domains are unknown: select or construct environments with higher state information richness and planning complexity, enable explicit reasoning, and apply light state randomization to improve generalization.
}

\date{\today}
\correspondence{\email{zhihanliu2027@u.northwestern.com}, \email{linguan@meta.com}, \email{nazhang@meta.com}}

\begin{document}

\maketitle

\newcommand{\idcell}[1]{\cellcolor{gray!15}\text{#1}}
\section{Introduction}
\label{sec:intro}

Post-training for \emph{generalist} large language model (LLM) agents is typically carried out on a narrow slice of tasks, while deployment spans far broader and shifting environments. 
Even with well-maintained data flywheels, the full cycle of sampling, post-processing, and running large-scale training remains compute and time intensive. As a result, post-training can only incorporate a limited set of domains and tasks \citep{luo2024arena}. Empirically, this skew shows up when agents that look strong on popular leaderboards only deliver subpar user experience: recent studies find sizable gaps between benchmarked strengths and performance in real-world ``office'' tasks \citep{xu2024agentcompany,miller2025realworld,jimenez2023swebench,jain2024livecodebench}. 
Therefore, when selecting or constructing post-training datasets, a key question is what types of task the agent should be trained on to ensure that the model not only achieves strong in-domain performance but also maintains generalization in unseen and broader scenarios.

On the other hand, online RL has become a principled tool for post-training LLMs, but it introduces its own constraints. Typical online RL training demands interactive environments that have reward signals and support the sampling of multiple rollouts. These simulation environments are costly to build and even infeasible for proprietary domains \citep{cao2024llmrlsurvey,wang2024rllmsurvey,samsami2024worldmodels}. The state of public agentic environments reflects this difficulty: despite notable progress, most benchmarks remain narrow (e.g., specific websites, platforms, or workflows) and expose only slices of the broader task distribution \citep{deng2023mind2web,zhou2023webarena,drouin2024workarena,boisvert2024workarenapp,lesellier2024browsergym,wang2024officebench}. Furthermore, even for production models deployed in real traffic, the resulting flywheel data is predominantly single-rollout and off-policy with respect to the updated policy in RL. Therefore, we aim to understand what characteristics make a training environment effective in preserving broader out-of-domain performance, so that we can prioritize constructing such environments and tasks for LLM agent post-training~\citep{guan2023leveraging, hu2025agentgen}. 

 


Overall, given the limited coverage in post-training data and the practical difficulty of constructing simulators at scale for online RL, we offer a data-centric perspective to study:
\begin{center}
\emph{How can we strategically sample tasks or selectively build domain simulators for RL training that maximally preserves or transfers agentic capabilities of LLMs to unseen domains? }
\end{center}

Before moving forward, we clarify that ``generalization'' has been assessed at different levels in prior work \citep{fu2025agentrefine,huan2025does,zhang2025agent}. The first is in-domain: models are tested on held-out cases where the task space is similar but not identical to that seen in training (e.g., shopping on Amazon vs. eBay). The second is cross-domain generalization: transferring to tasks with substantially different tools, states, or goals (e.g., from shopping to mathematical problem solving). Much of the reported evidence for ``RL generalization'' concerns the first type, while our interest here lies in the second. In our preliminary experiments on multiple agentic domains, as RL training proceeds, performance on held-out (unseen) environments drops while in-domain scores improve. Therefore, this work is more prevention-focused: identify training- and generation-time choices that maximally preserve performance in environments not observed during post-training.

By performing RL training on four distinct environments (Table \ref{tab:env_comparison}), namely Webshop \citep{yao2022webshop}, Sokoban \citep{junghanns2001sokoban}, ALFWorld \citep{Shridhar2021Alfworld}, and SciWorld \citep{Wang2022ScienceWorld}, we find that two axes exhibit high correlation with \emph{cross-domain performance}:

\begin{itemize}
\item \emph{State information richness}: the volume of information that the agent must perceive and process from the state. Higher richness increases the \emph{perception load}, forcing the policy to actively extract task-relevant signals from dense or noisy inputs. 
\item \emph{Planning complexity}: the difficulty of composing long and contingent action sequences, which imposes a higher \emph{reasoning load}. We estimate this via goal reachability under a base policy. Greater complexity exercises decomposition, progression, and potentially correction over a long horizon.
\end{itemize}

\begin{table}[htbp]
\centering
\footnotesize
\caption{Comparison of four agentic environments studied in this paper.}
\label{tab:env_comparison}
\begin{tabular}{p{1.6cm} p{5.0cm} p{2.6cm} p{2.6cm} p{2.6cm}}
\toprule
\textbf{Environment} & \textbf{Task Description} & \textbf{State Information Richness} & \textbf{Planning \,\,\,Complexity} & \textbf{Cross-Domain \,\,\,\,\,\,\,\,\,\,\,\,\,\, Robustness} \\
\midrule
\textbf{WebShop} & Web interaction to find the appropriate item for purchase & \emph{High} (3063) & \emph{Low} (33.9) & \emph{Medium} \\
\addlinespace[3pt]
\textbf{Sokoban} & Grid puzzle: push a single box to the target without getting stuck & \emph{High} (3114) & \emph{High} (44.0) & \emph{High} \\
\addlinespace[3pt]
\textbf{ALFWorld} & Household tasks across multiple rooms and objects & \emph{Low} (1572) & \emph{Medium} (42.0) & \emph{Low} \\
\addlinespace[3pt]
\textbf{SciWorld} & 
Similar manipulations tasks as in ALFWorld but performed in labs & \emph{Medium} (2851) & \emph{High} (43.5) & \emph{High} \\
\bottomrule
\end{tabular}
\end{table}

Given that state information richness and planning complexity are key factors for generalization, one natural follow-up question is whether we can make training environments more effective by manipulating either of these properties. In this work, we begin with state information richness as it is not intrinsic to a domain. Specifically, we apply state randomization by injecting small amounts of goal-irrelevant or domain-irrelevant content into the agent’s observations during training. This increases the information richness of states without altering other components in task space, and consistently improves cross-domain generalization. These results not only validate our hypothesis but also provide a practical method for augmenting any existing training environments. 

Alongside the main experiments, we also examine two modeling choices relevant to generalization. First, we find that mid-training has a strong and lasting influence on generalization. Specifically, it deeply infuses knowledge into the model, such that performance on covered domains remains stable even after subsequent RL training on other domains. However, the performance drops significantly on domains not included in mid-training, revealing a clear trade-off: while mid-training consolidates knowledge and improves robustness to downstream training, it risks erasing prior knowledge in uncovered domains. Moreover, we find that enabling step-by-step thinking during both training and evaluation does not improve in-domain scores, but plays a key role in preserving generalization. 
Taken together, all these findings offer concrete guidance for post-training when the target environment is unknown: select or construct training environments that have greater state information richness and planning complexity, enable explicit reasoning, and apply lightweight state randomization to better preserve generalization.


\medskip
\noindent\textbf{Contributions.} Our main contributions in this work are:

\begin{enumerate}
\item \textbf{Systematic investigation.} We systematically study cross-domain generalization in agentic RL and provide the first attempt to uncover its underlying environmental factors, showing that state information richness and planning complexity are important drivers of transferability to unseen environments.
\item \textbf{Causal validation and practical solutions.} We intervene on state information richness via a state randomization scheme that injects small amounts of goal- or domain-irrelevant content into state observations, consistently improving cross-domain robustness and providing causal validation of our findings. We further examine the effects of mid-training and explicit step-by-step thinking, uncovering their roles in generalization and the associated trade-offs.
\end{enumerate}

\section{Related Work}
\subsection{Multi-turn RL Training for LLM Agents} 

Recent works tackle the challenges of training LLMs as interactive agents via RL in multi-turn settings \citep{yao2022react,sun2023adaplanner,liu2023reason,wang2025ragen}. For example, \cite{wang2025ragen} and \cite{Feng2025verl-agent} extend the popular RL training framework VeRL \citep{sheng2024hybridflow} from the standard single-turn settings (as in coding and math tasks) to the multi-turn settings. Besides, 
\citet{zhang2025agentrl} introduce a scalable framework for agentic multi-turn RL that features an asynchronous generation-training pipeline and \citet{xi2025agentgym} propose a unified platform for training LLM agents in diverse environments. \citet{zhou2025sweet} focus on the credit assignment problem in long dialogues. They develop a benchmark for collaborative multi-turn tasks such as team-based software development. These advances in LLM-agent frameworks, together with prior works such as AlphaGo \citep{silver2016mastering}, demonstrate that with well-designed training environments and optimization strategies, RL agents can acquire highly specialized knowledge. Our work further expands the perspective of training specialized agents by investigating what types of environments are most effective for training generalist RL agents.

 \subsection{Cross-Domain Analysis for LLM Training} 
 As LLMs are applied to a wide range of tasks, recent studies have examined how well RL fine-tuning generalizes to different domains. \citet{cheng2025cross} empirically measure the transferability of skills acquired in a training domain to other unseen domains. 
Besides, \citet{huan2025does} compare the cross-domain generalization between SFT and RL on single-turn tasks and find that RL preserves more generalization than SFT.
These findings highlight the importance of understanding how an RL-tuned LLM performs on tasks beyond its training domains. Our work aligns with these perspectives, but with a particular emphasis on multi-turn agentic tasks, which are not well covered in previous studies.
Moreover, previous analyses largely focus on quantifying the correlation between training domains and performance on held-out domains. Our work moves beyond this by dissecting these correlations and explaining them through relevant environmental factors and modeling choices, which further informs the derivation of practical solutions for better generalization.

\subsection{State Augmentation for Generalization and Robustness}
Domain and state augmentation have long been explored as strategies to improve generalization, particularly in sim-to-real settings where agents trained in simulation must transfer to real-world deployments. These methods typically inject variations into the training environment by modifying background textures, object properties, lighting, or physics parameters \citep{tobin2017domain,sadeghi2017cad2rl,muratore2018domain,loquercio2019deep}.
More recent work has applied similar ideas in visual RL by augmenting observations or introducing distractors to improve robustness across domains or simulator variants \citep{pinto2017robustness,packer2018assessing,hansen2021soft,guan2021widening,yue2019domain,chen2021understanding,wang2025comprehensive}. It is important to note that our contribution lies beyond the technique of state randomization itself; we utilize it essentially as a tool to validate our systematic findings. Nevertheless, adapting it to LLM agent training is also meaningful: our work is the first to demonstrate semantic textual noise and controlled distraction degree are helpful for preventing OOD performance degradation in agentic RL.

\section{Preliminary}

\subsection{RL Training for LLM Agents}

\textbf{Environment and Agent.}
We consider a reinforcement learning framework in which a language model functions as an agent interacting with an environment to solve multi-step tasks described by a textual prompt \(x \sim p(X)\).
At each discrete timestep \(t \in \{1, \ldots, T\}\), the agent perceives a state \(s_t \in \mathcal{S}\) and produces a textual action \(a_t \in \mathcal{V}^n\), where \(\mathcal{V}\) represents the token set and \(n\) is the maximum number of tokens generated.
The environment then emits a scalar feedback \(r_t \in \mathbb{R}\) and transitions to a subsequent state \(s_{t+1}\).
An entire episode can be represented by the trajectory
$
\tau = \{(s_1, a_1, r_1), (s_2, a_2, r_2), \ldots, (s_T, a_T, r_T)\}.
$
The agent’s behavior is dictated by a conditional policy \(\pi_\theta(a_t \mid s_t, x)\) parameterized by \(\theta\), which defines the distribution over textual outputs given the current state and task prompt. The goal of the LLM agent is to find a policy $\pi$ that maximizes the corresponding expected cumulative reward $\mathbb{E}[R(\tau) \mid \tau \sim \pi]$, where $R(\tau) = \sum_{t=1}^T r_t$ denotes the sum of rewards along the trajectory, and the expectation is taken with respect to the stochastic policy $\pi$. 
In many practical agentic cases, the environment provides only binary success indicators at the end of an episode. 
As a result, the cumulative reward $R(\tau) $ is also sparse, which enables direct adoption of single-turn optimization approaches like GRPO \citep{shao2024deepseekmath}.

\textbf{Reinforcement Learning for LLM Agents.}
Recent advancements in large-scale language model training have popularized a \emph{group-based} strategy for reinforcement learning.
For a specific task prompt \(x\), the agent samples \(N\) full-episode trajectories \(\{\tau_1, \tau_2, \ldots, \tau_N\}\) with the latest policy.
Each trajectory \(\tau_i\) receives a scalar score \(R(\tau_i)\), representing the overall quality of the episode outcome.
Rather than estimating per-token value functions as in traditional actor–critic methods, group-based algorithms such as GRPO \citep{shao2024deepseekmath} derive normalized advantages using only the statistics computed from this group of trajectories:
\[
A(a_t^{(i)}) = \frac{R(\tau_j) - \text{Mean}(\{R(\tau_j)\}_{j=1}^{N})}{\text{Std}(\{R(\tau_j)\}_{j=1}^{N})} ,
\]
for all actions $\{a_t^{(i)}\}_{t=1}^T $ that belong to a trajectory $\tau_j$.
And the RL optimization objective for GRPO is:
\begin{align*}
\mathcal{J}_{\mathrm{GRPO}}(\theta)
&=
\mathbb{E}_{x \sim p(X)}
\,
\mathbb{E}_{\{\tau_i\}_{i=1}^{N} \sim \pi_{\theta_{\text{old}}}}
\left[
\frac{1}{NT}
\sum_{i=1}^{N}
\sum_{t=1}^{T}
\min\!\left(
\rho_\theta(a_t^{(i)})A(a_t^{(i)}),
\,
\mathrm{clip }\,\big(\rho_\theta(a_t^{(i)}), 1 - \epsilon, 1+ \epsilon\big)A(a_t^{(i)})
\right)
\right] \\&\qquad
-
\beta\,\mathbb{D}_{\mathrm{KL}}\!\big(
\pi_\theta(\cdot \mid x)
\parallel
\pi_{\mathrm{ref}}(\cdot \mid x)
\big),
\end{align*}

\noindent
where the importance sampling ratio between the current policy $\pi_\theta$ and the sampling policy $\pi_{\text{old}}$ is given by
\begin{equation*}
\rho_\theta(a_t^{(i)}) =
\frac{
\pi_\theta(a_t^{(i)} \mid s_t^{(i)}, x)
}{
\pi_{\theta_{\text{old}}}(a_t^{(i)} \mid s_t^{(i)}, x)
}.
\end{equation*}

\noindent
Similar to many multi-turn RL works \citep{Feng2025verl-agent,zhang2025agentrl}, GRPO in its vanilla form can be seamlessly extended to multi-turn tasks with sparse rewards.

\subsection{Cross-Domain Generalization of RL Training}\label{sec:cross-domain}
In this subsection, we provide a definition of cross-domain generalization and clarify its connection to other types of generalization.  
Let $e_i$ denote a text-based agentic environment, and let $\mathcal{T}_i$ be the set of tasks associated with $e_i$.  
The most basic form of generalization evaluation tests a model’s performance on unseen tasks \emph{within the same environment}.  
Formally, for each environment $e_i$, we partition its task set into two non-overlapping subsets,
$
\mathcal{T}_i \;=\; \mathcal{T}_i^{\text{train}} \,\cup\, \mathcal{T}_i^{\text{val}}, 
\qquad 
\mathcal{T}_i^{\text{train}} \,\cap\, \mathcal{T}_i^{\text{val}} \;=\; \emptyset.
$
Let $\mathcal{I}_{\text{env}}$ be the index set of all environments. We then define the training and evaluation task collections
$
\mathcal{E}^{\text{train}}_{\text{ID}} \;=\; \bigcup_{i \in \mathcal{I}_{\text{env}}} \mathcal{T}_i^{\text{train}},
\qquad
\mathcal{E}^{\text{eval}}_{\text{ID}} \;=\; \bigcup_{i \in \mathcal{I}_{\text{env}}} \mathcal{T}_i^{\text{val}}.
$
We refer to this setup as \emph{in-domain (ID) validation}.

Beyond this, one can evaluate a generalist agent on \emph{unseen environments} that may differ substantially in states, actions, and dynamics.  
Let $\mathcal{I}_{\text{env}}^{\text{train}}$ and $\mathcal{I}_{\text{env}}^{\text{eval}}$ denote the index sets of environments used for training and evaluation, respectively, with
$
\mathcal{I}_{\text{env}}^{\text{train}} \,\cap\, \mathcal{I}_{\text{env}}^{\text{eval}} \;=\; \emptyset.
$
We define the corresponding sets of environments
$
\mathcal{E}^{\text{train}} \;=\; \{\, e_i \mid i \in \mathcal{I}_{\text{env}}^{\text{train}} \,\},
\qquad
\mathcal{E}^{\text{eval}} \;=\; \{\, e_i \mid i \in \mathcal{I}_{\text{env}}^{\text{eval}} \,\}.
$
We refer to this setup as the evaluation of \emph{cross-domain generalization} or \emph{out-of-domain (OOD) generalization}.
Due to the limited availability of public agentic domains, for cross-domain evaluation, we use one environment for training at a time and use the remaining for evaluation.


\section{Analytical Framework}
\label{sec:setup}

\subsection{Domains and Metrics}
\label{sec:domain}

We conduct our analysis on four representative agentic environments~\citep{wang2025ragen,zhang2025agent}: WebShop~\citep{yao2022webshop}, Sokoban~\citep{junghanns2001sokoban}, ALFWorld~\citep{Shridhar2021Alfworld}, and SciWorld~\citep{Wang2022ScienceWorld}. 
These domains cover distinct types of common agentic tasks, ranging from web navigation and information retrieval to spatial reasoning and embodied interaction. 
While these environments vary across multiple dimensions—such as domain realism, action space size, and task horizon—our study identifies \emph{state information richness} and \emph{planning complexity} as the two primary factors influencing generalization. We will provide concrete definitions and measurements for these two axes in Section~\ref{sec:analysis} to quantitatively analyze their correlation with out-of-domain performance.

\begin{itemize}
    \item \textbf{WebShop}~\citep{yao2022webshop}: A text-based e-commerce simulator requiring the agent to navigate pages, filter attributes, and select products matching complex user queries. 
    
    \item \textbf{Sokoban}~\citep{junghanns2001sokoban}: A classic grid-based puzzle where the agent needs to push boxes to targets while avoiding irreversible lock-states. Unlike the matrix representation used in prior work~\citep{wang2025ragen}, we use explicit coordinate representations of walls, players, objects, and the target as the state representation. We find this necessary because RL training with \texttt{Llama-8B-instruct} fails to yield meaningful performance gains when using the original matrix representation \citep{wang2025ragen}.   
    
    \item \textbf{ALFWorld}~\citep{Shridhar2021Alfworld}: A widely adopted benchmark that requires the agent to generate symbolic actions (e.g., ``open fridge'', ``take apple'') to interact with a household environment.
    
    \item \textbf{SciWorld}~\citep{Wang2022ScienceWorld}: An interactive simulator of a scientific laboratory. We include this as a significantly more challenging extension of ALFWorld. SciWorld features a larger action space and the requirements for agents to manipulate specialized tools and materials to execute complex, multi-stage experiments. Examples that compare between ALFWorld and SciWorld are provided in Appendix~\ref{app:example}.
\end{itemize}

\noindent\textbf{Evaluation Protocol.} 
We perform RL training on one source domain and evaluate the performance on the remaining three to assess OOD generalization. Training runs for 150 steps, and we save checkpoints every 15 steps. We report the mean success rate (pass@1 per task) averaged over the last four checkpoints (i.e., checkpoints from steps 105 to 150). All experiments are repeated three times with different seeds.

\subsection{Models}
\label{sec:model}

\linchange{We adopt \texttt{Llama-3.1-8B-Instruct}~\citep{Grattafiori2024Llama3} as the base model. To establish a viable starting point for analyzing out-of-domain performance, the initial policy must exhibit non-trivial performance on all evaluation domains such that we can observe how the success rate changes over the course of subsequent RL on other training domains. However, the base model has a near-zero success rate in two domains, i.e., WebShop and SciWorld.} 

\linchange{For WebShop, conducting a 20-step RL training on it can increase the in-domain success rate to 34.4\%. The resulting policy, Checkpoint V1 ($\texttt{Ckpt V1}$), serves as an initial policy for all downstream experiments to be presented. However, for SciWorld, both the base model and $\texttt{Ckpt V1}$ remain at a near-zero success rate even after extensive RL training. As a result, we resort to warming up the base model using SFT with expert demonstrations. This setup is similar to mid-training ~\citep{wang2025octothinker,mo2025mid}, which imparts sufficient prior knowledge for effective subsequent RL training. To ensure balanced initial performance on all domains, we also mixed self-generated trajectories from ALFWorld and WebShop into the SFT dataset. The details of the SFT dataset and training can be found in Appendix \ref{app:domains}. We refer to this mid-training checkpoint as Checkpoint V2 ($\texttt{Ckpt V2}$).}

\linchange{In summary, $\texttt{Ckpt V1}$ and $\texttt{Ckpt V2}$ serve as the two initial policies for our downstream experiments, with $\texttt{Ckpt V2}$ additionally enabling us to investigate the influence of SFT warm-up or mid-training.}


\section{Characterizing Effective Training Domains for Preserving Generalization}
\label{sec:analysis}

\begin{table}[htbp]
\centering
\footnotesize
\setlength{\tabcolsep}{10pt}
\caption{Out-of-domain (OOD) success rates after RL training. \linchangenew{Shaded cells correspond to in-domain (ID) evaluation. To align with the discussion in Section~\ref{sec:analysis}, we report the change in ID success rate relative to the corresponding base model (i.e., $\Delta_{\text{ID}}$)}. Note that SciWorld evaluations for \texttt{Ckpt V1} are omitted as the success rates remain near zero as explained in Section~\ref{sec:model}.}
\label{tab:baseline}
\begin{tabular}{@{}cc*{4}{c}@{}}
\toprule
\multirow{2}{*}{\textbf{Initial Policy}} & \multirow{2}{*}{\textbf{Training Domain}} &
\multicolumn{4}{c}{\textbf{Evaluation Domain}} \\
\cmidrule(lr){3-6}
& & ALFWorld & Webshop & Sokoban & SciWorld \\
\midrule
\multirow{4}{*}{\texttt{Ckpt V1}} & Base Model & 25.8 & 34.4 & 12.5 & 3.1 \\
\cmidrule(lr){2-6}
& ALFWorld & \idcell{$\Delta_{\text{ID}}$=47.2} & 15.2 & 14.0 & -- \\
& Webshop & 10.8 & \idcell{$\Delta_{\text{ID}}$=23.6} & 12.8 & -- \\
& Sokoban & 20.5 & 15.0 & \idcell{$\Delta_{\text{ID}}$=36.0} & -- \\
\midrule
\multirow{5}{*}{\texttt{Ckpt V2}} & Base Model & 21.9 & 38.3 & 14.1 & 18.0 \\
\cmidrule(lr){2-6}
& ALFWorld & \idcell{$\Delta_{\text{ID}}$=60.6} & 30.5 & 9.8 & 10.0 \\
& Webshop & 17.0 & \idcell{$\Delta_{\text{ID}}$=28.7} & 9.0 & 13.8 \\
& Sokoban & 20.0 & 34.0 & \idcell{$\Delta_{\text{ID}}$=35.7} & 13.0 \\
& SciWorld & 19.8 & 35.8 & 12.0 & \idcell{$\Delta_{\text{ID}}$=20.2} \\
\bottomrule
\end{tabular}
\end{table}


\begin{table}[htbp]
\centering
\footnotesize
\caption{Correlation between properties of training environments and OOD robustness. \textbf{Avg. Character Count} denotes the average character count of states, which is used as an approximate measure for state information richness. \textbf{Avg. Trajectory Length} denotes the average trajectory length, which indicates planning horizon and complexity. \textbf{OOD Ranking Score} ($\downarrow$) represents cross-domain robustness, calculated by summing a model's rank (based on success rate) among all other evaluated checkpoints for each OOD domain. A lower OOD Ranking Score corresponds to a better overall rank, indicating stronger generalization.}
\label{tab:env_correlation}
\begin{tabular}{lcccccc}
\toprule
\multirow{2}{*}{\textbf{Training Domain}} & \multirow{2}{*}{\textbf{Avg. Character Count }} & \multirow{2}{*}{\textbf{Avg. Trajectory Length}} & \multicolumn{2}{c}{\textbf{OOD Ranking Score} ($\downarrow$)} \\\cmidrule(lr){4-5}
&&  & \texttt{Ckpt V1} and \texttt{V2} & \texttt{Ckpt V2}\\
\midrule
\textbf{SciWorld} & Medium (2851) & High (43.5) & -- & 3 (1st) \\
\textbf{Sokoban}  & High (3114)   & High (44.0) & 7 (1st) & 5 (2nd)\\
\textbf{WebShop}  & High (3063)   & Low (33.9)  & 10 (2nd) & 6 (3rd)\\
\textbf{ALFWorld} & Low (1572)    & Medium (42.0) & 10 (2nd) &  8 (4th)\\
\bottomrule
\end{tabular}
\end{table}

This section presents the initial experiment results based on the setup outlined in Section~\ref{sec:setup}. We aim to identify the properties of training environments that best align with the preservation of agent capabilities in unseen domains. We first report the performance of agents trained on each domain and evaluated on all others in Table~\ref{tab:baseline}. Consistent with prior studiess~\citep{luo2024arena}, while RL training significantly improves in-domain performance, it often leads to performance drops in OOD environments.

\linchangenew{For a concise and clear comparison of how different training domains affect robustness, we introduce a ranking-based metric. For each training domain, we calculate its performance rank on every OOD domain (e.g., 1 represents the best performance compared against checkpoints trained on other training domains) and compute the sum of these ranks. We refer to this sum as OOD Ranking Score.} 

\linchangenew{The aggregated results are shown in Table~\ref{tab:baseline}, with detailed ranks provided in Table~\ref{tab:rank} in Appendix. Recall from Section~\ref{sec:model} that $\texttt{Ckpt V1}$ and $\texttt{Ckpt V2}$ are applicable to different training domains due to SciWorld's requirement for SFT warmup. Consequently, we report two sets of aggregated results: one based solely on Ckpt V2, and the other combining Ckpt V1 and Ckpt V2. The former compares all training domains (including SciWorld), while the latter compares the other three domains across a broader set of experimental trials. With \texttt{Ckpt V2}, SciWorld yields the most robust training outcome (ranking score: 3), closely followed by Sokoban (ranking score: 5). WebShop ranks third (ranking score: 6), while ALFWorld performs the worst (ranking score: 8). A similar pattern is observed in the combined results (\texttt{Ckpt V1} and \texttt{Ckpt V2}), where Sokoban is more effective (ranking score: 7) than both WebShop and ALFWorld, which are tied with a ranking score of 10.}

\linchangenew{Based on these rankings, our goal is to identify factors that correlate with stronger OOD performance. Given the complexity and the exploratory nature of such an analysis, we do not aim to exhaustively identify all factors, but rather to reveal at least some key insights that can inform the selection or construction of training domains.}
Overall, we find two planning-related concepts below better explain the effectiveness of environments. For a quantitative analysis, we ground these two concepts into two quantifiable metrics, which are formally defined and measured as follows:

\begin{enumerate}
    \item \textbf{State Information Richness:} State information richness refers to the volume of information contained in the environment states. A higher state information richness means that the agent must process a larger amount of content from the state to extract useful information for the current decision-making. We note that the observations in all four studied domains are highly compact, which convey each piece of state information in a similarly concise manner (see Appendix~\ref{app:example} for examples). In this case, a feasible way of quantifying information richness would be to  use the \emph{average character count of environment states}. 
    Practically, we employ \texttt{Ckpt V2} to collect 128 trajectories per domain with a maximum step limit $T_{\max}=50$ and calculate the average length of the states within the sampled trajectories.
    \item \textbf{Planning Complexity:} Planning complexity refers to the difficulty of composing a multi-step plan to solve the task. A higher planning complexity indicates that the task requires a longer, more sustained planning chain and has a lower goal reachability. Although planning complexity is often measured by properties such as branching factors or search depth in the classical planning literature \citep{de2016planning}, these metrics are less applicable to LLM agents that leverage strong priors rather than relying on conventional search space traversal. Instead, we use \emph{average trajectory length} as an empirical proxy of search depth to measure the \textit{reasoning effort} in order to reach the terminal states.
    Specifically, we also employ \texttt{Ckpt V2} to collect 128 trajectories per domain with a maximum step limit $T_{\max}=50$ and compute this metric by averaging the trajectory lengths, where failed trajectories are assigned the maximum length $T_{\max}$ to reflect the inability to solve the task within the budget. The average trajectory length in this case captures both task horizon and goal reachability.
\end{enumerate}

Table~\ref{tab:env_correlation} shows how the state information richness and the planning complexity are correlated to the robustness rankings. Specifically, Sokoban exhibits high state richness (3114 characters) and high planning complexity (44.0 steps), while SciWorld similarly combines substantial planning complexity (43.5 steps) with moderate-to-high information richness. This indicates that environments that demand both high perception loads and deep reasoning chains foster better generalization preservation. 
In contrast, the two less effective domains, namely WebShop and AlfWorld, exhibit lower complexity in at least one dimension. Specifically, although WebShop has a richer state space characterized by lengthy product descriptions and feature lists, it demands relatively low planning complexity. A WebShop agent generally only needs to iterate through the products to identify a match based on attributes. Similarly, while AlfWorld involves more complicated task logic, its state space remains relatively simple and compact.


From Table~\ref{tab:baseline}, we also observe that another potentially correlated factor could be the change in ID success rates (i.e., $\Delta_{\text{ID}}$) over the course of RL. The results seem to suggest a naive trade-off: intuitively, the more specialized knowledge absorbed through RL training, the lower the model's OOD performance. This raises the question of whether knowledge absorption is a more dominant factor than the environment properties discussed above. To answer this, we conducted an additional experiment where we extended the training steps for Sokoban and SciWorld to further increase their $\Delta_{\text{ID}}$, comparing the final checkpoints to the ALFWorld and WebShop checkpoints at 150 steps. As shown in Table \ref{tab:additional_run}, even when the $\Delta_{\text{ID}}$ of Sokoban is increased to match or exceed that of ALFWorld and WebShop, its OOD performance remains higher. These results confirm the impact of the two environment properties and suggest that OOD performance is not simply determined by $\Delta_{\text{ID}}$. Additionally, we also identify several qualitative factors that have low correlations with OOD generalization. For instance, domain realism does not explain the performance differences. Sokoban, an abstract grid-world puzzle, enables strong generalization even to realistic domains like SciWorld. In contrast, ALFWorld, which simulates realistic household tasks, ranks lowest for cross-domain robustness.


\begin{table}[htbp]
\centering
\footnotesize
\setlength{\tabcolsep}{10pt}
\caption{OOD performance of Sokoban and SciWorld with extended training steps. The SciWorld checkpoints were trained with fewer extra steps due to earlier convergence. The full results, including OOD success rates and ranking scores, are provided in Table~\ref{tab:app_eval_cont_rl} in Appendix. Here, $\Delta_{\text{ID}}$ indicates the change in ID success rate relative to the corresponding base model.} 
\label{tab:additional_run}
\begin{tabular}{@{}ccccccc@{}}
\toprule
\multirow{2}{*}{\textbf{Training Domain}} &\multirow{2}{*}
{\textbf{Total Training Steps}} &
\multirow{2}{*}{\textbf{Initial Policy}} & 
\multirow{2}{*}{$\Delta_{\text{ID}}$} &
\multicolumn{2}{c}{\textbf{OOD Ranking Score} ($\downarrow$)}

 \\\cmidrule(lr){5-6}
& & &  & \texttt{Ckpt V1} and \texttt{V2}& \texttt{Ckpt V2} \\\midrule
\multirow{4}{*}{Sokoban}
& \multirow{2}{*}{150} & \texttt{Ckpt V1} & 36.0 & \multirow{2}{*}{7 (1st)} & \multirow{2}{*}{5 (2nd)} \\
& & \texttt{Ckpt V2} & 35.7 & &  \\\cmidrule(lr){2-6}
& \multirow{2}{*}{300}  & \texttt{Ckpt V1} & 67.9 &\multirow{2}{*}{8 (1st)} & \multirow{2}{*}{6 (2nd)} \\
& & \texttt{Ckpt V2} & 68.4 &  &  \\\midrule
\multirow{2}{*}{SciWorld}
& 150 & {\texttt{Ckpt V2}} & 20.2 & -- & 3 (1st)
 \\\cmidrule(lr){2-6}
& 240 & {\texttt{Ckpt V2}}& 41.8 & -- & 5 (1st) \\
\bottomrule
\end{tabular}
\end{table}

As an implication, one could strategically select or synthesize training environments with greater state information richness and planning complexity to better preserve out-of-domain generalization. Nonetheless, given the limited number of available domains, our current evidence should be interpreted as indicative rather than conclusive. We acknowledge that the initial characterization of effective domains is relatively ad hoc and may not causally isolate the critical factors. Therefore, in the next section, we extend the analysis through additional experiments designed to test some of these hypotheses and provide practical approaches for better generalization.

\section{From Correlation to Practical Solution for Preserving Generalization}
\label{sec:randomization}

The analysis in Section~\ref{sec:analysis} identifies two environmental properties, namely planning complexity and state information richness, which align with cross-domain generalization. A natural next step is to intervene along either of these axes to test whether modifying them can causally enhance generalization. Since planning complexity is largely intrinsic to the task logic and difficult to manipulate without altering the environment's underlying dynamics, we focus on increasing \emph{state information richness} through a technique we call \emph{State Information Augmentation}. This approach involves enriching observations with goal-irrelevant information while keeping task goals, action spaces, and reward functions unchanged. Intuitively, this approach forces the agent to more actively extract task-relevant signals while ignoring distractive noise, which reduces the model's reliance on domain-specific patterns.

Formally, let $s$ denote the original state and $g$ the goal. For each state $s$, we sample a short goal-irrelevant text fragment $\delta$ from a distribution of content that does not interfere with the original optimal policy conditioned on $g$. We thereby construct an augmented state $s' = \mathrm{Augment}(s, \delta, \epsilon)$ by injecting a controlled volume $\epsilon$ of goal-irrelevant information $\delta$ into the textual state. Note that the augmentation is applied only to the agent's input; components such as transitions and rewards still operate on the original state.

Practically, there can be several options for coming up with valid augmentations that do not interfere with the original optimal plan. With minimal domain knowledge, one could inject observations from a dramatically different domain into the training environment. For instance, we can inject Webshop observation into an Alfworld state by appending ``there is a monitor showing a shopping webpage: <random webshop state>.'' With slightly more understanding of a domain, one can come up with more contextual-relevant augmentation. For example, for the ALFWorld task ``slice an apple'', valid augmentations could involve describing unrelated static objects (e.g., ``the wall is painted blue'') or redundant features (e.g., ``the apple is red''). Examples of augmentation for all training domains can be found in Appendix~\ref{app:aug}. Nevertheless, we note that this study emphasizes verifying how increased state information richness can facilitate cross-domain generalization, instead of focusing extensively on the exact technique for constructing state augmentation.

Another practical consideration is setting the volume ($\epsilon$) of distractive information, which can also be quantified using the character count of the injected text. Our experiments reveal insufficient augmentation may not yield a noticeable change in OOD performance, while an excessive amount of noise can make the tasks overly difficult for the agent to learn, thereby negatively affecting ID performance. In practice, the appropriate $\epsilon$ can be determined by interactively increasing its value until a significant drop in ID success rate is observed. Furthermore, we found it beneficial to stochastically apply the augmentation to only a portion of trajectories within a batch. This allows for better control over learning difficulty on top of adjusting the noise volume. In our experiments, we apply the augmentation to each trajectory with a probability of 50\%, and the results suggest that this strategy helps maintain ID performance.



\begin{table}[htbp]
\centering
\footnotesize
\setlength{\tabcolsep}{8pt}
\caption{Impact of state information augmentation on OOD success rates after RL training. Here, \textbf{+ State Info. Aug.} denotes the model trained with state information augmentation. Shaded cells correspond to in-domain (ID) evaluation. To align with the discussion in Section~\ref{sec:analysis}, we report the change in ID success rate relative to the corresponding base model (i.e., $\Delta_{\text{ID}}$). The \textbf{OOD Change} column reports the difference in OOD performance between the model using augmented states and the baseline using original states. A positive value indicates improvement.}
\label{tab:aug}
\begin{tabular}{@{}cc*{5}{c}@{}}
\toprule
\multirow{2}{*}{\textbf{Initial Policy}} & \multirow{2}{*}{\textbf{Training Domain}} &
\multicolumn{4}{c}{\textbf{Evaluation Domain}} & \multirow{2}{*}{\textbf{OOD Change}} \\
\cmidrule(lr){3-6}
& & ALFWorld & Webshop & Sokoban & SciWorld & \\
\midrule
\multirow{7}{*}{\texttt{Ckpt V1}} & Base Model & 25.8 & 34.4 & 12.5 & 3.1 & -- \\
\cmidrule(lr){2-7}
& ALFWorld & \idcell{$\Delta_{\text{ID}}$=47.2} & 15.2 & 14.0 & -- & -- \\
& \,\, + State Info. Aug. & \idcell{$\Delta_{\text{ID}}$=47.7} & 23.8 & 11.2 & -- & $+32.6\%$ \\
\cmidrule(lr){2-7}
& Webshop & 10.8 & \idcell{$\Delta_{\text{ID}}$=23.6} & 12.8 & -- & -- \\
& \,\, + State Info. Aug. & 17.8 & \idcell{$\Delta_{\text{ID}}$=27.6} & 11.0 & -- & $+35.5\%$ \\
\cmidrule(lr){2-7}
& Sokoban & 20.5 & 15.0 & \idcell{$\Delta_{\text{ID}}$=36.0} & -- & -- \\
& \,\, + State Info. Aug. & 21.2 & 24.8 & \idcell{$\Delta_{\text{ID}}$=36.7} & -- & $+42.5\%$ \\
\midrule
\multirow{7}{*}{\texttt{Ckpt V2}} & Base Model & 21.9 & 38.3 & 14.1 & 18.0 & -- \\
\cmidrule(lr){2-7}
& ALFWorld & \idcell{$\Delta_{\text{ID}}$=60.6} & 30.5 & 9.8 & 10.0 & -- \\
& \,\, + State Info. Aug. & \idcell{$\Delta_{\text{ID}}$=56.6} & 30.3 & 11.0 & 12.5 & $+7.0\%$ \\
\cmidrule(lr){2-7}
& Webshop & 17.0 & \idcell{$\Delta_{\text{ID}}$=28.7} & 9.0 & 13.8 & -- \\
& \,\, + State Info. Aug. & 25.8 & \idcell{$\Delta_{\text{ID}}$=27.7} & 11.8 & 15.5 & $+33.4\%$ \\ 
\cmidrule(lr){2-7}
& Sokoban & 20.0 & 34.0 & \idcell{$\Delta_{\text{ID}}$=35.7} & 13.0 & -- \\
& \,\, + State Info. Aug. & 20.8 & 37.0 & \idcell{$\Delta_{\text{ID}}$=37.4} & 13.0 & $+5.7\%$ \\
\bottomrule
\end{tabular}

\end{table}

Table~\ref{tab:aug} presents the empirical results of this intervention. To rigorously quantify the effect of state information augmentation, we introduce the \textbf{OOD Change} metric ($\Delta_{\text{OOD}}$). Let $\pi^{\text{RL}}_{\text{w/o Aug}}$ and $\pi^{\text{RL}}_{\text{w/ Aug}}$ denote the policies trained with the original states and augmented states, respectively. Following the notations in Section~\ref{sec:cross-domain}, let $\mathcal{E}^{\text{eval}}$ be the set of unseen evaluation environments for a given training session, and $G(\pi, e)$ be the success rate of policy $\pi$ on environment $e$. We define the aggregate OOD change as:
{\small
\[
\Delta_{\text{OOD}} = \frac{\sum_{e \in \mathcal{E}^{\text{eval}}} \left( G(\pi^{\text{RL}}_{\text{w/ Aug}}, e) - G(\pi^{\text{RL}}_{\text{w/o Aug}}, e) \right)}{\sum_{e \in \mathcal{E}^{\text{eval}}} G(\pi^{\text{RL}}_{\text{w/o Aug}}, e)} \times 100\%.
\]
}

A positive $\Delta_{\text{OOD}}$ indicates that state information augmentation effectively improves the agent's generalization. For most combinations of training domains and base policies, state information augmentation yields positive $\Delta_{\text{OOD}}$. In the experiments with \texttt{Ckpt V1}, where augmentation universally improves OOD performance across all training domains: ALFWorld ($+7.0\%$), WebShop ($+35.5\%$), and Sokoban ($+42.5\%$). We observe a similar positive trend in the \texttt{Ckpt V2} setting for ALFWorld ($+7.0\%$), WebShop ($+33.4\%$), and Sokoban ($+5.7\%$). These results causally validate the hypothesis that state information richness is a determinant of agent robustness. Moreover, the technique of state information augmentation can also serve as a low-cost and practical solution for preserving cross-domain capabilities.



\section{Other Influential Factors on Cross-domain Generalization}
\label{sec:design}

\linchange{During the experiments on the environment-side intervention, we also determined two modeling-side design choices that have a significant influence on OOD performance. This section proceeds by first discussing how SFT warmup impacts knowledge retention and forgetting during the subsequent RL phase, followed by an analysis of the necessity of explicit reasoning for both ID and OOD performance.}

\subsection{SFT Warmup Reshapes Knowledge Retention and Forgetting of Subsequent RL}
\label{sec:midtraining}

\linchange{Recall from Section~\ref{sec:model} that \texttt{Ckpt V2} is obtained via SFT warmup to ensure a viable starting point for OOD analysis. This specific setup further provides an unique setup to study how SFT warmup reshapes retention and forgetting by comparing $\texttt{Ckpt V2}$ and $\texttt{Ckpt V1}$, which does not undergo extra SFT training. Specifically, we compare the OOD performance of these two checkpoints after identical downstream RL training. Our results (Table~\ref{tab:sft_comparison}) indicate that SFT warmup has its distinct advantages and disadvantages from the perspective of generalization.}

\begin{table}[htbp]
\centering
\footnotesize
\setlength{\tabcolsep}{5pt}\caption{Impact of SFT warmup on OOD success rates (\%) after RL training. We compare OOD performance of models initialized with \texttt{Ckpt V1} against \texttt{Ckpt V2} after identical downstream RL training. \textbf{Rel. Change} indicates the percentage change in success rate relative to the corresponding initial policy. \textbf{Rel. Change Diff. (Ckpt V1 to V2)} shows the difference of the Rel. Change of Ckpt V2 and the  Rel. Change of Ckpt V1. \linchange{The column} ``\textbf{In V2 SFT-warmup datamix?}'' indicates whether the evaluation domain \linchange{is covered} in the SFT warmup datamix of \texttt{Ckpt V2}.}
\label{tab:sft_comparison}
\begin{tabular}{@{}llcccccc@{}}
\toprule
\multirow{3}{*}{\textbf{\shortstack{RL Training\\Domain}}} & \multirow{3}{*}{\textbf{Eval Domain}} & \multirow{2}{*}{\textbf{\shortstack{In V2\\SFT-Warmup\\ Datamix?}}} & \multicolumn{2}{c}{\textbf{Ckpt V1}} & \multicolumn{2}{c}{\textbf{Ckpt V2}}&\multirow{2}{*}{\textbf{Rel. Change Diff. (Ckpt V1 to V2)}} \\
\cmidrule(lr){4-5} \cmidrule(lr){6-7}
& & & \textbf{success rate} & \textbf{Rel. Change} & \textbf{success rate} & \textbf{Rel. Change} &\\
\midrule
\multirow{2}{*}{\textbf{Sokoban}} & WebShop & Yes & 15.0 & $-56.4\%$ & 34.0 & $\mathbf{-11.2\%}$ & $+45.2\%$ \\
& ALFWorld & Yes & 20.5 & $-20.5\%$ & 20.0 & $\mathbf{-8.7\%}$ & $+11.8\%$ \\
\midrule
\multirow{2}{*}{\textbf{ALFWorld}} & WebShop & Yes & 10.8 & $-68.6\%$ & 17.8 & $\mathbf{-53.5\%}$ & $+15.1\%$ \\
& Sokoban & No & 14.0 & $+\mathbf{12.0}\%$ & 9.8 & $-30.5\%$ & $-42.5\%$ \\
\midrule
\multirow{2}{*}{\textbf{WebShop}} & ALFWorld & Yes & 10.8 & $-58.1\%$ & 17.0 & $\mathbf{-22.4\%}$ & $+35.7\%$ \\
& Sokoban & No & 12.8 & $\mathbf{+2.4}\%$ & 9.0 & $-36.2\%$ & $-38.6\%$ \\
\bottomrule
\end{tabular}
\end{table}

\linchange{On the positive side, SFT warmup provides a strong lever for making domain knowledge more robust to downstream RL training. As shown in the ``In V2 SFT-warmup datamix” column in Table~\ref{tab:sft_comparison}, the SFT-warmup datamix of \texttt{Ckpt V2} consists of data from three domains: ALFWorld, WebShop, and SciWorld. While SFT warmup does not change the fact that OOD performance declines throughout RL training, this decline is consistently and substantially smaller on evaluation domains covered by SFT warmup. For example, with Sokoban as the RL training domain and WebShop as the evaluation domain, the success rate of \texttt{Ckpt V1} drops by $-56.4\%$ whereas that of \texttt{Ckpt V2} only drops by $-11.2\%$. Likewise, with WebShop as the RL training domain and ALFWorld as the evaluation domain, the success rate of \texttt{Ckpt V1} drops by $-58.1\%$ whereas that of \texttt{Ckpt V2} only drops by $-22.4\%$. The same pattern holds in all experiments with evaluation domains covered by SFT warmup. Moreover, we note that the warmup data for ALFWorld and WebShop is directly sampled from the base model itself without extra filtering (e.g., with success indicator). This suggests that simply exposing the model to environment dynamics during SFT warmup can effectively consolidate domain knowledge.}


\linchange{Nevertheless, SFT warmup also sacrifices performance on domains outside its coverage, making knowledge from these domains more vulnerable to forgetting. This is evident when evaluating on Sokoban where \texttt{Ckpt V2} consistently exhibits worse performance than \texttt{Ckpt V1}. For instance, when training on WebShop, \texttt{Ckpt V1}'s OOD performance on Sokoban slightly increases by $+2.4\%$ whereas that of \texttt{Ckpt V2} decreases by a large margin ($-36.2\%$).}

\linchange{These results suggest an inherent trade-off. SFT warmup or mid-training can effectively infuse and consolidate knowledge within a policy model, but it also carries a significant risk of erasing or diminishing prior capabilities in domains that are not covered. When the deployment environment is unknown, it is crucial to carefully control the extent of mid-training and to ensure that the datamix is as broad as possible to mitigate the risk of catastrophic forgetting.}


\subsection{Enabling Step-by-Step Reasoning Preserves OOD Performance}
\label{sec:reasoning}

\linchange{Another important modeling-side consideration is the use of explicit step-by-step reasoning during RL training. Although explicit reasoning has become the default strategy for many LLM-driven agentic systems, it is worth noting that ``reactive'' architectures, such as DQN \citep{mnih2015human, silver2016mastering}, can also achieve human-level performance on target domains without explicit reasoning. Here, we provide an additional perspective: explicit reasoning is essential for transferring or preserving agentic capabilities in novel domains that are not seen during training.}


\begin{table}[htbp]
\centering
\footnotesize
\setlength{\tabcolsep}{8pt}
\caption{Impact of Step-by-Step Reasoning on OOD success rates (\%)  after RL training.  We compare the baseline RL runs (which use thinking) against runs where thinking is disabled (\textbf{w/o Thinking}). Shaded cells correspond to in-domain (ID) evaluation. To align with the discussion in Section~\ref{sec:analysis}, we report the change in ID success rate relative to the base model (i.e., $\Delta_{\text{ID}}$). The \textbf{OOD Change} column reports the relative performance drop of the non-thinking setup compared to the baseline. A large negative value indicates that disabling thinking severely hurts generalization.}
\label{tab:nothink}
\begin{tabular}{@{}cc*{5}{c}@{}}
\toprule
\multirow{2}{*}{\textbf{Initial Policy}} & \multirow{2}{*}{\textbf{Training Domain}} &
\multicolumn{4}{c}{\textbf{Evaluation Domain}} & \multirow{2}{*}{\textbf{OOD Change}} \\
\cmidrule(lr){3-6}
& & ALFWorld & Webshop & Sokoban & SciWorld & \\
\midrule
\multirow{7}{*}{\texttt{Ckpt V1}} & Base Model & 25.8 & 34.4 & 12.5 & 3.1 & -- \\
\cmidrule(lr){2-7}
& ALFWorld & \idcell{$\Delta_{\text{ID}}$=47.2} & 15.2 & 14.0 & -- & -- \\
& \,\, w/o Thinking & \idcell{$\Delta_{\text{ID}}$=51.2} & 0.8 & 7.5 & -- & $-119.0\%$ \\
\cmidrule(lr){2-7}
& Webshop & 10.8 & \idcell{$\Delta_{\text{ID}}$=23.6} & 12.8 & -- & -- \\
& \,\, w/o Thinking & 8.2 & \idcell{$\Delta_{\text{ID}}$=27.6} & 17.8 & -- & $+16.9\%$ \\
\cmidrule(lr){2-7}
& Sokoban & 20.5 & 15.0 & \idcell{$\Delta_{\text{ID}}$=36.0} & -- & -- \\
& \,\, w/o Thinking & 16.0 & 17.2 & \idcell{$\Delta_{\text{ID}}$=23.0} & -- & $-9.1\%$ \\
\midrule
\multirow{9}{*}{\texttt{Ckpt V2}} & Base Model & 21.9 & 38.3 & 14.1 & 18.0 & -- \\
\cmidrule(lr){2-7}
& ALFWorld & \idcell{$\Delta_{\text{ID}}$=60.6} & 30.5 & 9.8 & 10.0 & -- \\
& \,\, w/o Thinking & \idcell{$\Delta_{\text{ID}}$=51.3} & 1.0 & 6.5 & -- & $-234.2\%$ \\
\cmidrule(lr){2-7}
& Webshop & 17.0 & \idcell{$\Delta_{\text{ID}}$=28.7} & 9.0 & 13.8 & -- \\
& \,\, w/o Thinking & 4.5 & \idcell{$\Delta_{\text{ID}}$=14.5} & 12.5 & 11.0 & $-82.5\%$ \\
\cmidrule(lr){2-7}
& Sokoban & 20.0 & 34.0 & \idcell{$\Delta_{\text{ID}}$=35.7} & 13.0 & -- \\
& \,\, w/o Thinking & 9.2 & 25.8 & \idcell{$\Delta_{\text{ID}}$=60.7} & 7.8 & $-216.5\%$ \\
\cmidrule(lr){2-7}
& SciWorld & 19.8 & 35.8 & 12.0 & \idcell{$\Delta_{\text{ID}}$=20.2} & -- \\
& \,\, w/o Thinking & 6.2 & 32.5 & 14.0 & \idcell{$\Delta_{\text{ID}}$=17.5} & $-216.9\%$ \\
\bottomrule
\end{tabular}
\end{table}

\linchange{Table~\ref{tab:nothink} demonstrates that models trained with explicit reasoning consistently maintain higher OOD performance compared to those reactive baselines that output actions directly. In experiments with \texttt{Ckpt V2}, disabling reasoning causes the OOD performance to collapse by over 200\% in the ALFWorld, Sokoban, and SciWorld training runs. For instance, when training on ALFWorld and evaluating on WebShop, the success rate of the policy trained without explicit reasoning drops to 1.0\%, whereas the policy with explicit reasoning maintains a success rate of 30.5\%.}

Notably, enabling reasoning does not always benefit in-domain performance. In several cases, such as WebShop and ALFWorld, policies trained without thinking achieve comparable or even slightly higher in-domain success rates. For instance, the in-domain performance on ALFWorld using \texttt{Ckpt V1} increases from 73.0\% to 77.0\% when reasoning is disabled. This suggests \linchange{reactive policies can sufficiently learn the optimal policies by discovering domain-specific patterns tailored for the training environments. However, these patterns or heuristics can be brittle and not transferrable. By forcing the model to articulate its reasoning process step-by-step, the model is less prone to overfitting to shallow heuristics and thereby preserving more cross-domain generalization.}


\section{Limitations and Future Work}
\label{sec:limits}

This work establishes a practical framework for analyzing and improving cross-domain generalization in agentic post-training. Looking ahead, several promising directions merit further exploration.
First, expanding this study to a broader suite of domains and diverse model scales would help validate the universality of factors identified as critical for OOD performance. A larger-scale investigation may also reveal additional factors that influence generalization. 
Second, there is room to further refine the empirical measures used to characterize effective training environments. Developing more formal and causally grounded definitions would support more principled environment design.
Methodology-wise, future work could investigate automated environment augmentation techniques, the relationship between reasoning structure and generalization, and adaptive SFT warmup schedules that dynamically balance knowledge consolidation and flexibility. 
Lastly, connecting these empirical findings to theoretical interpretations of generalization and policy optimization remains an important step toward establishing quantitative design principles for scalable and robust agent training.

\section{Conclusion}
\label{sec:conclusion}

In this work, we take a data-centric perspective on RL post-training for \emph{generalist} LLM agents, and probe how to strategically select or construct training domains so that agentic capabilities can be maximally preserved in unseen domains. 
By analyzing correlations between cross-domain performance and properties of training environments, we find that realism and surface-level semantic similarity are not the main factors. Instead, \emph{state information richness} and \emph{planning complexity}, which respectively control perception load and reasoning load, align more with OOD performance.

Building upon these observations, we causally validate the importance of state information richness by directly intervening on this dimension. We introduce a state randomization method that injects small amounts of goal-irrelevant content into the agent’s observations, increasing information volume without changing the task space. Beyond environment-side designs, we also examine two important modeling choices, namely mid-training and step-by-step thinking. Our results reveal that mid-training or SFT warmup can deeply infuse domain knowledge into the policy model, and the infused knowledge can retain even after subsequent RL that is conducted on other domains. However, mid-training can also lead to larger generalization degradation on uncovered domains compared to RL. We further find that, while enabling step-by-step thinking during RL and evaluation does not increase ID success rates, it plays a crucial role in transferring agentic capabilities to unseen environments. Taken together, these results provide practical guidance for post-training when future test environments are unknown: select or construct training environments with higher state information richness and planning complexity, enable explicit step-by-step thinking, and apply lightweight state randomization to better maintain robustness and generalization.

\clearpage
\newpage
\bibliographystyle{assets/plainnat}
\bibliography{paper}
\definecolor{seedblue}{HTML}{2E5AA8}
\tcbset{
  dialogbox/.style={
    colback=seedblue!5,
    colframe=seedblue!60,
    boxrule=0.5pt,
    arc=2mm,
    outer arc=2mm,
    left=2mm,
    right=2mm,
    top=1mm,
    bottom=1mm,
    fonttitle=\bfseries,
    before skip=5pt,
    after skip=5pt,
  }
}

\lstdefinestyle{prompt}{
    basicstyle=\ttfamily\fontsize{9pt}{9pt}\selectfont,
    frame=none,
    breaklines=true,
    backgroundcolor=\color{seedblue!10},
    breakatwhitespace=false,
    columns=fullflexible,
    breakindent=0pt,
    escapeinside={(*@}{@*)},
    numbers=none,
    numbersep=5pt,
    xleftmargin=5pt,
}

\clearpage
\newpage
\beginappendix






\begin{table}[htbp]
\centering
\footnotesize
\setlength{\tabcolsep}{10pt}
\caption{Performance ranking of models trained on different source domains when evaluated on each unseen domain. For each evaluation column, the training domains are ranked from best (1) to worst (3 or 4) based on their success rate. Domains with success rate differences $<$ 0.5 are considered tied and assigned the same rank. In-domain entries are marked with --.}
\label{tab:rank}
\begin{tabular}{@{}cc*{4}{c}@{}}
\toprule
\multirow{2}{*}{\textbf{Initial Policy}} & \multirow{2}{*}{\textbf{Training Domain}} &
\multicolumn{4}{c}{\textbf{Evaluation Domain}} \\
\cmidrule(lr){3-6}
& & ALFWorld & Webshop & Sokoban & SciWorld \\
\midrule
\multirow{3}{*}{\texttt{Ckpt V1}} & ALFWorld & -- & 1 & 1 & -- \\
& Webshop & 2 & -- & 2 & -- \\
& Sokoban & 1 & 1 & -- & -- \\
\midrule
\multirow{4}{*}{\texttt{Ckpt V2}} & ALFWorld & -- & 3 & 2 & 3 \\
& Webshop & 2 & -- & 3 & 1 \\
& Sokoban & 1 & 2 & -- & 2 \\
& SciWorld & 1 & 1 & 1 & -- \\
\bottomrule
\end{tabular}
\end{table}


\begin{table}[htbp]
\centering
\footnotesize
\setlength{\tabcolsep}{10pt}
\caption{OOD success rates of Sokoban and SciWorld with extended training steps.}
\label{tab:app_eval_cont_rl}
\begin{tabular}{@{}ccccccc@{}}
\toprule
\multirow{2}{*}{\textbf{Training Domain}} &\multirow{2}{*}
{\textbf{Total Training Steps}} &
\multirow{2}{*}{\textbf{Initial Policy}} & 
\multicolumn{4}{c}{\textbf{Evaluation Domain}}
 \\\cmidrule(lr){4-7}
& &  & ALFWorld & Webshop & Sokoban & SciWorld\\
\midrule
\multirow{4}{*}{Sokoban}
& \multirow{2}{*}{150} & \texttt{Ckpt V1} & 20.5 & 15.0  &  \idcell{$\Delta_{\text{ID}}$=36.0}& --\\
 & &\texttt{Ckpt V2} & 20.0 &  34.0 & \idcell{$\Delta_{\text{ID}}$=35.7} & 13.0 \\\cmidrule(lr){2-7}
& \multirow{2}{*}{300}  & \texttt{Ckpt V1} & 19.2 & 15.8 &  \idcell{$\Delta_{\text{ID}}$=67.9}& -- \\
& & \texttt{Ckpt V2} & 17.5 &  36.9 &    \idcell{$\Delta_{\text{ID}}$=68.4}&  9.7 \\\midrule
\multirow{2}{*}{SciWorld}
& 150 & {\texttt{Ckpt V2}} & 19.8  & 35.8 & 12.0 & \idcell{$\Delta_{\text{ID}}$=20.2}
 \\\cmidrule(lr){2-7}
& 240 & {\texttt{Ckpt V2}}& 16.9 &  33.6 &  11.8 & \idcell{$\Delta_{\text{ID}}$=41.8} \\
\bottomrule
\end{tabular}
\end{table}

\begin{table}[htbp]
\centering
\footnotesize
\setlength{\tabcolsep}{10pt}
\caption{OOD performance of Sokoban and SciWorld with extended training steps. The SciWorld checkpoints were trained with fewer extra steps due to earlier convergence. Domains with success rate differences $<$ 0.5 are considered tied and assigned the same rank.}
\label{tab:app_eval_cont_rl}
\begin{tabular}{@{}ccccccc@{}}
\toprule
\multirow{2}{*}{\textbf{Training Domain}} &\multirow{2}{*}
{\textbf{Total Training Steps}} &
\multirow{2}{*}{\textbf{Initial Policy}} & 
\multicolumn{4}{c}{\textbf{Evaluation Domain}}
 \\\cmidrule(lr){4-7}
& &  & ALFWorld & Webshop & Sokoban & SciWorld\\
\midrule
\multirow{4}{*}{Sokoban}
& \multirow{2}{*}{150} & \texttt{Ckpt V1} & 1 &  1 &  \idcell{$\Delta_{\text{ID}}$=36.0}& --\\
 & &\texttt{Ckpt V2} & 1 & 2& \idcell{$\Delta_{\text{ID}}$=35.7} &  2\\\cmidrule(lr){2-7}
& \multirow{2}{*}{300}  & \texttt{Ckpt V1} & 1& 1&  \idcell{$\Delta_{\text{ID}}$=64.5}& -- \\
& & \texttt{Ckpt V2} &  2 &  1 &  \idcell{$\Delta_{\text{ID}}$=66.5}& 3  \\\midrule
\multirow{2}{*}{SciWorld}
& 150 & {\texttt{Ckpt V2}} &1  &  1& 1& \idcell{$\Delta_{\text{ID}}$=20.2}
 \\\cmidrule(lr){2-7}
& 240 & {\texttt{Ckpt V2}}& 2 & 2& 1 & \idcell{$\Delta_{\text{ID}}$=40.7} \\
\bottomrule
\end{tabular}
\end{table}

\section{Experimental Details}
\label{app:domains}

\subsection{RL in ALFWorld}
We use verl-agent \citep{Feng2025verl-agent} as the codebase and use 8 NVIDIA A100 (80G) GPUs for the training and evaluation of ALFWorld. We perform the evaluation using the out-of-domain split of the ALFWorld dataset and perform the training using the train split.

\textbf{Hyperparameters for ALFWorld.} The maximum prompt length is 2048 tokens, and the maximum response length is 512 tokens. Each
episode allows up to 50 environment steps. The learning rate is set to 1e-6 for the actor. We adopt a rule-based reward, assigning a reward of 10 for success
and 0 for failure. To handle invalid actions generated by the agent, we apply a reward penalty of
-0.1. For GRPO, we sample 8 trajectories per task prompts (i.e., group size = 8) and use 16 task prompts per training step.   The rollout temperature is set to 1.0, while the validation temperature is set
to 0.4. The training mini-batch size consists of 256 (representing the state count), and the KL-divergence regularizer coefficient is set to 0.01.
\subsection{RL in Webshop}
We use verl-agent \citep{Feng2025verl-agent} as the codebase and use 8 NVIDIA A100 (80G) GPUs for the training and evaluation of Webshop. We use the first 500 goals from the ``small dataset'' with 1000 products for evaluation and the remaining ``small dataset'' for training.

\textbf{Hyperparameters for Webshop.} The maximum prompt length is 4096 tokens, and the maximum response length is 1024 tokens. Each
episode allows up to 15 environment steps. The learning rate is set to 1e-6 for the actor. We adopt a rule-based reward, assigning a reward of 10 for success
and 0 for failure. To handle invalid actions generated by the agent, we apply a reward penalty of
-0.1. For GRPO, we sample 8 trajectories per task prompts (i.e., group size = 8) and use 16 task prompts per training step.   The rollout temperature is set to 1.0, while the validation temperature is set
to 0.4. The training mini-batch size is 256 (representing the state count), and the KL-divergence regularizer coefficient is set to 0.01.

\subsection{RL in Sokoban}
We use verl-agent \citep{Feng2025verl-agent} as the codebase and use 8 NVIDIA A100 (80G) GPUs for the training and evaluation of Sokoban. 

\textbf{Hyperparameters for Sokoban.} The maximum prompt length is 4096 tokens, and the maximum response length is 512 tokens. Each
episode allows up to 15 environment steps. The learning rate is set to 1e-6 for the actor. We adopt a rule-based reward, assigning a reward of 10 for success
and 0 for failure. To handle invalid actions generated by the agent, we apply a reward penalty of
-0.1. For GRPO, we sample 8 trajectories per task prompts (i.e., group size = 8) and use 16 task prompts per training step.   The rollout temperature is set to 1.0, while the validation temperature is set
to 0.4. The training mini-batch size is 256 (representing the state count), and the KL-divergence regularizer coefficient is set to 0.01.

\subsection{RL in SciWorld}
Since the SciWorld environment is incompatible with the verl-agent, we modify verl-agent \citep{Feng2025verl-agent} to implement the interaction interface via a proxy-client framework to decouple agent interaction, LLM inference, and model training.  
We use 8 NVIDIA A100 (80G) GPUs for the training and evaluation of SciWorld.

\textbf{Hyperparameters for SciWorld} The maximum prompt length is 2048 tokens, and the maximum response length is 1024 tokens. Each
episode allows up to 50 environment steps. The learning rate is set to 1e-6 for the actor. We adopt a rule-based reward, assigning a reward of 10 for success
and 0 for failure. To handle invalid actions generated by the agent, we apply a reward penalty of
-0.1. For GRPO, we sample 8 trajectories per task prompts (i.e., group size = 8) and use 16 task prompts per training step.   The rollout temperature is set to 1.0, while the validation temperature is set
to 0.4. The training mini-batch size is 256 (representing the state count), and the KL-divergence regularizer coefficient is set to 0.01.
\subsection{SFT Warmup}
We use LlamaFactory \citep{zheng2024llamafactory} as the codebase for SFT warmup. The SFT dataset is constructed by mixing expert SciWorld data \citep{Wang2022ScienceWorld}, ALFWorld data generated by \texttt{Ckpt V1}, and WebShop data generated by \texttt{Ckpt V1}  in a 19:3:1 ratio. This ratio is defined over the count of state-action pairs.

\textbf{Hyperparameters for SFT Warmup} 
The learning rate is 5e-6 with the cosine learning rate scheduler and warmup ratio 0.1. The batch size is 32. We train the model for 100 steps to best balance the performance on SciWorld and the performance on ALFWorld, Webshop, and Sokoban.

\section{Concrete Examples}
\label{app:example}
In this section, we provide example states and actions from ALFWorld, Webshop, Sokoban, and SciWorld for comparison.
\subsection{ALFWorld}
\textbf{Example State from ALFWorld}
\begin{lstlisting}[style=prompt]
You are an expert agent operating in the ALFRED Embodied Environment. Your current observation is: -= Welcome to TextWorld, ALFRED! =-

You are in the middle of a room. Looking quickly around you, you see a cabinet 16, a cabinet 15, a cabinet 14, a cabinet 13, a cabinet 12, a cabinet 11, a cabinet 10, a cabinet 9, a cabinet 8, a cabinet 7, a cabinet 6, a cabinet 5, a cabinet 4, a cabinet 3, a cabinet 2, a cabinet 1, a coffeemachine 1, a countertop 2, a countertop 1, a diningtable 1, a drawer 5, a drawer 4, a drawer 3, a drawer 2, a drawer 1, a fridge 1, a garbagecan 1, a microwave 1, a safe 1, a sinkbasin 1, a stoveburner 4, a stoveburner 3, a stoveburner 2, a stoveburner 1, and a toaster 1.

Your task is to: find two knife and put them in drawer. Your admissible actions of the current situation are: ['go to cabinet 1' 'go to cabinet 10' 'go to cabinet 11' 'go to cabinet 12' 'go to cabinet 13' 'go to cabinet 14' 'go to cabinet 15' 'go to cabinet 16' 'go to cabinet 2' 'go to cabinet 3' 'go to cabinet 4' 'go to cabinet 5' 'go to cabinet 6' 'go to cabinet 7' 'go to cabinet 8' 'go to cabinet 9' 'go to coffeemachine 1' 'go to countertop 1' 'go to countertop 2' 'go to diningtable 1' 'go to drawer 1' 'go to drawer 2' 'go to drawer 3' 'go to drawer 4' 'go to drawer 5' 'go to fridge 1' 'go to garbagecan 1' 'go to microwave 1' 'go to safe 1' 'go to sinkbasin 1' 'go to stoveburner 1' 'go to stoveburner 2' 'go to stoveburner 3' 'go to stoveburner 4' 'go to toaster 1' 'inventory' 'look'].

Now it's your turn to take an action. You should first reason step-by-step about the current situation. This reasoning process MUST be enclosed within <think> </think> tags.  Once you've finished your reasoning, you should choose an admissible action for current step and present it within <action> </action> tags.

\end{lstlisting}

\textbf{Example Action (with Thinking) from ALFWorld}

\begin{lstlisting}[style=prompt]
<think> The current situation is that I am in the middle of a room, and I need to find two knives and put them in a drawer. I have a list of admissible actions that I can take to achieve this goal. The first step is to determine which drawer I should go to. Since I need to put the knives in a drawer, I should go to the drawer that is closest to the knives. I have already observed the room, and I see that there are several drawers, but I don't know which ones contain knives. I need to go to the drawer that is most likely to contain knives.

The most likely drawer to contain knives is the one that is closest to the kitchen area, where the knives are typically kept. I see that there are several cabinets and appliances in the room, but I don't see any knives. I need to go to the drawer that is closest to the kitchen area. The closest drawer to the kitchen area is drawer 5. </think> <action>go to drawer 5</action>



\end{lstlisting}

\subsection{Webshop}
\textbf{Example State from Webshop}
\begin{lstlisting}[style=prompt]
You are an expert autonomous agent operating in the WebShop e-commerce environment. Your task is to: Find me slim fit, loose fit men's tuxedo shirts with long sleeve, short sleeve, contrast color, classic fit for teen girls with color: b-bk1, and size: small, and price lower than 30.00 dollars. Prior to this step, you have already taken 14 step(s). Below are the most recent 2 observations and the corresponding actions you took: [Observation 13: ''Back to Search' [SEP] '< Prev' [SEP] 'size' [SEP] 'medium' [SEP] 'large' [SEP] 'x-large' [SEP] 'xx-large' [SEP] '3x-large' [SEP] 'color' [SEP] 'white' [SEP] 'yellow' [SEP] 'Mens Linen Shirt,Men's Striped Shirts Casual Short Sleeve Button Down Shirts Regular Fit Hawaiian Shirts Beach Tees Tops' [SEP] 'Price: $3.78 to $11.38' [SEP] 'Rating: N.A.' [SEP] 'Description' [SEP] 'Features' [SEP] 'Reviews' [SEP] 'Buy Now'', Action 13: 'click[back to search]'] [Observation 14: ''Search'', Action 14: 'search[query: \"slim fit loose fit men's tuxedo shirts long sleeve short sleeve contrast color classic fit teen girls b-bk1 small lower than 30.00\"]'] You are now at step 15 and your current observation is: 'Back to Search' [SEP] 'Page 1 (Total results: 50)' [SEP] 'Next >' [SEP] 'B09S3BN15C' [SEP] 'Mens Linen Shirt,Men's Striped Shirts Casual Short Sleeve Button Down Shirts Regular Fit Hawaiian Shirts Beach Tees Tops' [SEP] '$3.78 to $11.38' [SEP] 'B09NNMV9LN' [SEP] 'JSPOYOU Mens Short Sleeve Crewneck 3D Graphic Tunic Shirts Big & Tall Tie Dye Summer Top Basic Designed Classic Cotton Shirt' [SEP] '$3.99 to $10.99' [SEP] 'B09Q67H373' [SEP] 'JSPOYOU 2022 Newly T-Shirt for Mens Funny 3D Graphics Pattern Crewneck Short Sleeve Tees Big and Tall Summer Casual Comfy Top' [SEP] '$1.99 to $8.99' [SEP] 'B09QQP3356' [SEP] 'HAUKLIE Men's Sports Waffle Ribbed Polo Shirts Summer Short Sleeve Cotton Muscle Quarter-Zip Henley T-Shirt Tunics Tops' [SEP] '$10.99' [SEP] 'B07F2G93BJ' [SEP] 'Amazon Essentials Men's Slim-fit Long-Sleeve Solid Pocket Oxford Shirt' [SEP] '$18.5' [SEP] 'B09QGK5XHZ' [SEP] 'WENKOMG1 Men's Long Sleeve Undershirt with Mask Turtleneck Hooded T-Shirt Solid Color Workout Tops Zipper Side Slit Shirts Slim Fit Sweatshirt Spring/Summer Tee Shirts(Gray,)' [SEP] '$8.39' [SEP] 'B09HGFV91W' [SEP] 'ZHDD Tops for Mens, Men's Knight Punk Vintage Style Long Sleeve Asymmetric Hem Cowl Neck Teen Boys Gothic Pullover' [SEP] '$4.98 to $7.99' [SEP] 'B09R9YCM6R' [SEP] 'XXBR Summer T-shirts for Mens, Soldier Short Sleeve 3D Street Vintage Printed Shirt Slim Fit Muscle Casual Tee Tops' [SEP] '$8.98 to $11.99' [SEP] 'B09T756KQ5' [SEP] 'WENKOMG1 Men's Crewneck Goth Tee Shirts Skull Print Tops Spring/Summer Long Sleeve Sports T-Shirt Baggy Y2K Soft Streetwear' [SEP] '$4.09 to $8.09' [SEP] 'B09ND9DP7J' [SEP] 'InterestPrint Gold Horse Pattern Men's 2-Piece Sleepwear Set, Long Sleeve Shirt with Pants Loungewear' [SEP] '$43.59'. Your admissible actions of the current situation are: [ 'click[back to search]', 'click[next >]', 'click[b09s3bn15c]', 'click[b09nnmv9ln]', 'click[b09q67h373]', 'click[b09qqp3356]', 'click[b09qqp3356]', 'click[b07f2g93bj], 'click[b09qgk5xhz]', 'click[b09hgfv91w]', 'click[b09r9ycm6r]', 'click[b09t756kq5]', 'click[b09nd9dp7j]', ].

Note: Prioritize click a product in the current page over going to next page. Now it's your turn to take one action for the current step (respond in English). You should first reason step-by-step about the current situation. This reasoning process MUST be enclosed within <think> </think> tags.  Once you've finished your reasoning, you should choose an admissible action for current step and present it within <action> </action> tags.

\end{lstlisting}

\textbf{Example Action (with Thinking) from Webshop}

\begin{lstlisting}[style=prompt]
<think> The current situation is that I have already searched for slim fit loose fit men's tuxedo shirts long sleeve short sleeve contrast color classic fit teen girls b-bk1 small lower than 30.00. The search returned 50 results, but none of them match the specified criteria of having a long sleeve, short sleeve, contrast color, classic fit for teen girls with color: b-bk1, and size: small, and price lower than 30.00 dollars.

The current page displays a list of products that do not match the specified criteria. The first product is a men's linen shirt, which does not have a long sleeve, short sleeve, contrast color, classic fit for teen girls, or a color of b-bk1. The other products also do not match the specified criteria. I need to review the products on the current page to see if any of them match the specified criteria. If not, I will need to go to the next page to review the remaining products. However, I need to prioritize reviewing the products on the current page first, as it is more efficient than going to the next page.</think>
<action> click[b09s3bn15c] </action>
\end{lstlisting}

\subsection{Sokoban}
\textbf{Example State from Sokoban}
\begin{lstlisting}[style=prompt]
You are an expert agent operating in the Sokoban environment.

# Objective
Your objective is to push all boxes onto their corresponding target locations. Once every box is correctly placed on a target, the puzzle is solved and you win.

# Environment Setup
- The environment is a 6\u00d76 grid. - Each cell in the grid can contain one of the following objects: **player (you)**, **box**, **target**, **wall**, or be **empty**. - All objects occupy exactly one grid cell and are specified using their (row, column) coordinates, where (0, 0) is the top-left corner.

# Rules
1. **Movement and Constraints**   - You can move up, down, left, or right by one cell at a time.   - You **cannot move through walls**.   - You **cannot move onto or through boxes** unless you are pushing them.   - You **cannot pull boxes**, only push them.

2. **Pushing Boxes**   - To push a box, it must be directly adjacent to your current position in the direction you are moving.   - The cell immediately beyond the box (in the same direction) must be empty (i.e., not a wall, another box, or the boundary).   - When you push a box, both you and the box move one cell forward.

**Example:**   If you are at (2, 3), a box is at (2, 4), and (2, 5) is empty, then moving right pushes the box to (2, 5), and you move to (2, 4).

3. **Deadlocks and Planning**   - Avoid pushing boxes into corners or against walls where they cannot be moved again.

# Observation Format
You will be given a list of all objects and their coordinates, including:   - Player position   - Box positions   - Target positions   - Wall positions   All unspecified cells are empty.

# Current Step
Prior to this step, you have already taken 27 step(s). Below are the most recent 1 observaitons and the corresponding actions you took: [Text Observation 27:  Wall at (0, 0) Wall at (0, 1) Wall at (0, 2) Wall at (0, 3) Wall at (0, 4) Wall at (0, 5) Wall at (1, 0) Wall at (1, 5) Wall at (2, 0) Goal at (2, 1) Wall at (2, 5) Wall at (3, 0) Box at (3, 3) Wall at (3, 5) Wall at (4, 0) Wall at (4, 1) Wall at (4, 2) Player at (4, 3) Wall at (4, 4) Wall at (4, 5) Wall at (5, 0) Wall at (5, 1) Wall at (5, 2) Wall at (5, 3) Wall at (5, 4) Wall at (5, 5) Action 27: 'Still'] You are now at step 28 and your current observation is: Wall at (0, 0) Wall at (0, 1) Wall at (0, 2) Wall at (0, 3) Wall at (0, 4) Wall at (0, 5) Wall at (1, 0) Wall at (1, 5) Wall at (2, 0) Goal at (2, 1) Wall at (2, 5) Wall at (3, 0) Box at (3, 3) Wall at (3, 5) Wall at (4, 0) Wall at (4, 1) Wall at (4, 2) Player at (4, 3) Wall at (4, 4) Wall at (4, 5) Wall at (5, 0) Wall at (5, 1) Wall at (5, 2) Wall at (5, 3) Wall at (5, 4) Wall at (5, 5) Your admissible actions are [\"up\", \"down\", \"left\", \"right\"].

Now it's your turn to make a move (choose ONE action only for the current step). You should first reason step-by-step about the current situation. This reasoning process MUST be enclosed within <think> </think> tags.  Once you've finished your reasoning, you should choose an admissible action for current step and present it within <action> </action> tags.
\end{lstlisting}

\textbf{Example Action (with Thinking) from Sokoban}

\begin{lstlisting}[style=prompt]
<think>The current situation is that I am at (4, 3) and I have a box at (3, 3) in front of me. The goal is to push the box to (2, 1) where the target is located. To achieve this, I need to move the box to the target location. I can see that the cell immediately beyond the box (in the same direction) is empty. This means that I can push the box to the target location. The current admissible actions are [\"up\", \"down\", \"left\", \"right\"]. Since the box is in front of me, I need to move to the right to push the box to the target location. </think>
<action>right</action>

\end{lstlisting}
\subsection{SciWorld}
\textbf{Example State from SciWorld}
\begin{lstlisting}[style=prompt]
You are an expert agent operating in the SCIWorld Embodied Environment.  ## Begin of an Example: Your task is to boil water. For compounds without a boiling point, combusting the substance is also acceptable. First, focus on the substance. Then, take actions that will cause it to change its state of matter. This outside location is called the outside. Here you see: the ground an axe the agent a substance called wood a fountain (containing nothing) a substance called air a fire pit (containing nothing) You also see: A door to the green house (that is open) A door to the foundry (that is open) A door to the kitchen (that is open)

<think> Going to the kitchen seems like a sensible choice because it is likely the place where I can find tools for boiling like a pot and a heat source. I am heading to the kitchen as it might contain equipment or a setup suitable for boiling water. </think><action>go kitchen</action> ## End of an Example

Now, your task is: Your task is to determine if unknown substance B is electrically conductive. The unknown substance B is located around the workshop. First, focus on the unknown substance B. If it is electrically conductive, place it in the blue box. If it is electrically nonconductive, place it in the orange box.. Prior to this step, you have already taken 7 step(s). Below are the most recent 2 observaitons and the corresponding actions you took: [Observation 6: 'You move the unknown substance B to the workshop.', Action 6: 'drop unknown substance in blue box'] [Observation 7: 'No known action matches that input.', Action 7: 'look around'] You are now at step 8 and your current observation is: This room is called the workshop. In it, you see:  the agent a substance called air a blue box (containing nothing) a green light bulb, which is off a orange box (containing nothing) a table. On the table is: a battery, a green wire, a orange wire, a red light bulb, which is off, a red wire, a switch, which is off, a violet light bulb, which is off, a yellow light bulb, which is off. a ultra low temperature freezer. The ultra low temperature freezer door is closed.  unknown substance B You also see: A door to the hallway (that is open) Your admissible actions of the current situation are: ['activate OBJ' 'close OBJ' 'connect OBJ to OBJ' 'deactivate OBJ' 'disconnect OBJ' 'dunk OBJ in OBJ' 'eat OBJ' 'flush OBJ' 'focus on OBJ' 'go OBJ' 'inventory' 'look around' 'look at OBJ' 'look in OBJ' 'mix OBJ' 'move OBJ to OBJ' 'open OBJ' 'pick up OBJ' 'pour OBJ in OBJ' 'put down OBJ' 'read OBJ' 'use OBJ on OBJ' 'wait'], where you should replace the 'OBJ' with the specific objects from the admissible objects list: ['agent' 'air' 'anode in battery' 'anode in component' 'anode in green light bulb' 'anode in red light bulb' 'anode in violet light bulb' 'anode in yellow light bulb' 'battery' 'battery cathode' 'blue box' 'cathode in component' 'cathode in green light bulb' 'cathode in red light bulb' 'cathode in violet light bulb' 'cathode in yellow light bulb' 'door' 'freezer' 'green light bulb' 'green wire' 'green wire terminal 1' 'green wire terminal 2' 'hallway' 'orange box' 'orange wire' 'orange wire terminal 1' 'orange wire terminal 2' 'red light bulb' 'red wire' 'red wire terminal 1' 'red wire terminal 2' 'switch' 'table' 'unknown substance' 'violet light bulb' 'workshop' 'yellow light bulb'].

Now it's your turn to take an action. Do remeber that do NOT repeat the same action in your previous steps such as 'look around' and try to explore different actions to achieve the goal. You should first reason step-by-step about the current situation. This reasoning process MUST be enclosed within <think> </think> tags.  Once you've finished your reasoning, you should choose an admissible action for current step and present it within <action> </action> tags.

\end{lstlisting}


\textbf{Example Action (with Thinking) from SciWorld}

\begin{lstlisting}[style=prompt]
<think> The goal is to determine if unknown substance B is electrically conductive. The unknown substance B is located around the workshop. If it is electrically conductive, place it in the blue box. If it is electrically nonconductive, place it in the orange box. The current situation is that I have already moved the unknown substance B to the workshop, and I am now in the workshop. The unknown substance B is currently in the blue box, which means that I have already placed it there. The current observation is that the unknown substance B is in the blue box, and I need to determine if it is electrically conductive. To do this, I need to check if the green light bulb is on when the switch is activated. If it is, then the unknown substance B is electrically conductive. If not, then it is electrically nonconductive. The current action is to drop the unknown substance B in the blue box, which means that I need to activate the switch to check if the green light bulb is on. </think><action>activate switch</action>
\end{lstlisting}

\section{Information Augmentation Details}
\label{app:aug}
In this section, we provide the implementation details of state augmentation. We also cover the selection of information volume for ALFWorld, Webshop, and Sokoban.

\subsection{ALFWorld}
For ALFWorld, we increase state information richness by introducing synthetic distractor objects. To introduce distractor objects that do not interfere with the original tasks, we employ a multi-step synthesis process. First, we sample an object type from a predefined list: \texttt{\{bowl, cup, pan, spoon, fork, knife, plate, mug, bottle, can, glass, basket, tray, vase, jar, container, jug, thermos, pitcher, dish, tumbler, sponge, towel, napkin, rack, coaster, utensil, lid\}}. Adhering to ALFWorld's naming convention of (\texttt{type, ID}), if the sampled object type currently exists in the scene, we will assign a distinct ID to ensure uniqueness. 

Finally, we sample a ``trivial'' object descriptor from the list: \texttt{\{cracked, dirty, slightly burnt, covered in dust, wrapped in foil, sticky, still warm, rusty, filled with water, emits a faint smell, has some liquid inside, tipped over, seems unused, greasy, wet on the surface, has a faint label, smudged, with faded color, shows fingerprints\}}. The objects and their descriptors are combined using short natural-language templates such as ``You notice a \ldots{} that looks \ldots'' or ``Near the sink, there is a \ldots{} that is \ldots''. Such distractive sentences are inserted at random positions in the textual states (i.e., observations). 

The augmentation strength is controlled by an information volume parameter \(\epsilon\). We estimate the length of each sentence as 12 tokens and inject \(n_{\text{distractive}} = \lfloor \epsilon / 12 \rfloor\) sentences per state. For example, \(\epsilon=120\) yields about 10 extra sentences, while \(\epsilon=360\) yields about 30. 

\subsection{Webshop}
We consider injecting distractive information into two types of pages in WebShop, namely product-details pages and search-result pages.

On both page types, we inject irrelevant advertising information in the form of (\texttt{clothing categories}, \texttt{promotional descriptors}). For clothing categories, we sample from the following list: \texttt{\{rompers, jumpsuits, blouses, cozy cardigans, pajamas, loungewear sets, button-down tops, long sleeve tees, casual jackets, knitwear, soft sweaters, thermal layers, fleece pullovers, v-neck tunics, red outfits, plus-size picks, cropped hoodies, layered looks, textured knits, everyday basics\}}. For promotional descriptors, we sample from the following list: \texttt{\{30\% off, new arrivals, trending now, selling fast, limited edition, back in stock, editor's pick, members-only deal, must-have looks, staff favorite, top-rated, recommended for daily wear, cozy \& cute, perfect for layering, style spotlight, hand-picked for you, fan favorites, most loved by shoppers, inspired by you, perfect for your capsule wardrobe\}}. Categories and descriptors are combined into short ad sentences via the template: \texttt{[AD] <descriptor> --- shop <category> today!}. 

On product-detail pages, we introduce trivial features sampled from the following list: \{\texttt{\{fabric: machine wash cold\}}, \texttt{\{shipping note: ships within 5-7 days\}}, \texttt{\{sizing tip: runs half size large\}}, \texttt{\{model info: model is 5'8'' wearing size medium\}}, \texttt{\{brand description: known for cozy minimalism\}}, \texttt{\{care label: made in Indonesia\}}, \texttt{\{origin: eco-conscious packaging\}}, \texttt{\{batch code: Spring 2024 release\}}, \texttt{\{warehouse tag: popular for travel \& errands\}}, \texttt{\{return policy: soft fleece interior\}}, \texttt{\{manufacturing month: delicate stitching on seams\}}, \texttt{\{recommendation: designed for colder climates\}}, \texttt{\{intended use: do not iron over print\}}, \texttt{\{layering advice: packaged folded in bag\}}\}.


On search-result pages, we also create trivial products by sampling non-targeted product types paired with non-targeted features. The product types include \texttt{\{romper, jumpsuit, blouse, tunic, pajama top, sweater, tank top, hoodie, cardigan, bodysuit, thermal shirt, dress, coverall, camisole\}}. Non-targeted features include \texttt{\{sleeveless, misleadingly tagged, overpriced, heavily discounted, out of stock, not available in large size, no reviews, size runs small, fabric mismatch, zipper broken, category mislabel, style mismatch, faded color, missing product images, customer complaints\}}. The information is presented with templates such as \texttt{Women <desc> <obj> --- likely unsuitable} or \texttt{Warning: <desc> <obj> (cannot be selected)}. 

The information volume \(\epsilon\) determines how many of such units are inserted. On each result page, we allocate up to 10 additional item slots and fill \(k = \min(\lfloor (\epsilon / 100)\,\alpha \cdot 10 \rfloor, 10)\) of them with a mix of advertisements and trivial products (e.g., with \(\epsilon=100\) and \(\alpha=0.5\), we insert \(k=5\) entries). On each detail page, we append \(k = \max(1, \lfloor \epsilon / 25 \rfloor)\) sentences about product features, and \(k = \max(1, \lfloor \epsilon / 30 \rfloor)\) sentences about advertisements. 


\subsection{Sokoban}
\textbf{Augmentation Implementation}
For Sokoban, the augmentation operates at the level of the textual description of the grid. Given the true room dimensions \(H \times W\), we sample coordinates that lie strictly outside the truly reachable locations and associate each with a random object drawn from the list: \texttt{\{Broken crate, Sealed door, Pile of rubble, Faded sign, Leaking pipe, Loose tile, Stacked boxe\}}. The locations are described using phrases from \texttt{\{outside the room, beyond the wall, inaccessible, behind a collapsed wall, on a blocked corridor\}}. 

These elements are combined with the template \texttt{(r, c) shows a <obj> (<desc>; unreachable).}. The final sentences are appended as additional lines after the original textual state. Here, the information volume \(\epsilon\) directly controls the number of appended sentences: we estimate the length of each sentence as 10 tokens and generate \(k = \max(1, \lfloor \epsilon / 10 \rfloor)\) distractor lines (e.g., \(\epsilon=50\) yields about 5 lines of unreachable locations, while \(\epsilon=150\) yields about 15). Again, all distractors are explicitly placed outside the grid.

\subsection{Information Augmentation Volume Selection}

In Tab. \ref{tab:info_volume}, we report the information augmentation volume $\epsilon$ used for the experiments in Section \ref{sec:randomization}.

\begin{table}[htbp]
\centering
\caption{Information augmentation volume $\epsilon$ used for the experiments in Section \ref{sec:randomization}.}
\label{tab:info_volume}
\begin{tabular}{@{}cccc@{}}
\toprule
\textbf{Initial Policy} & \textbf{Training Domain} & \textbf{Information Augmentation Volume $\epsilon$} & \textbf{Proability of Appying Augmentation} \\
\midrule
\texttt{Ckpt V1} & ALFWorld & 300 & 100\%  \\
\texttt{Ckpt V1} & Webshop  & 30 & 100\%  \\
\texttt{Ckpt V1} & Sokoban  & 80 & 100\%  \\
\hline
\texttt{Ckpt V2} & ALFWorld & 200 & 50\%  \\
\texttt{Ckpt V2} & Webshop  & 40 & 50\%  \\
\texttt{Ckpt V2} & Sokoban  & 80 & 50\%   \\
\bottomrule
\end{tabular}
\end{table}

\end{document}